\documentclass[subscriptcorrection,upint,varvw,barcolor=Goldenrod3,mathalfa=cal=euler,balance,hyphenate,french,pdf-a]{asmejour} %

\hypersetup{%
	pdfauthor={Xinyun Huo},                       		   	% <=== change to YOUR name[s]!
	pdftitle={MSEC Conference Paper},                  	% <=== change to YOUR pdf file title
	pdfkeywords={Large Language Model, Human-Robot Collaboration, Natural Language Understanding, Smart Manufacturing Systems, Structured Language Model},% <=== change to YOUR pdf keywords
	pdfsubject = {},			% <=== change to YOUR subject
	pdflicenseurl={},% may delete
}

\JourName{JCISE}%<=== change to the name of your journal

\begin{document}

% Change to your author name[s] and addresses, in the desired order of authors.
% First name, middle initial, last name
% Use title case (upper and lower case letters)
% Note usage below for corresponding author.

\SetAuthorBlock{Xinyun Huo}{Industrial and Manufacturing Engineering,\\
   Florida State University,\\
   Tallahassee, FL, USA \\
   email: xh25b@fsu.edu} 
\SetAuthorBlock{Raghav Gnanasambandam}{Industrial and Manufacturing Engineering,\\
   Florida State University,\\
   Tallahassee, FL, USA \\
   email: raghavg@eng.famu.fsu.edu} 

% To label one or more corresponding authors put "Name\CorrespondingAuthor". No space after "Name".
% An optional argument can be added if email is not in address block as
%      "Name\CorrespondingAuthor{write@to.me}"
% Can also include multiple emails and use the command more than once for multiple corresponding authors,
%      "Name\CorrespondingAuthor{write@to.him, write@to.her}"

\SetAuthorBlock{Xinyao Zhang\CorrespondingAuthor}{Industrial and Manufacturing Engineering,\\
   Florida State University,\\
   Tallahassee, FL, USA \\
   email: xz25f@fsu.edu} 

%%% Change to your paper title. Can insert line breaks if you wish (otherwise breaks are selected automatically).
\title{PRECISE ROBOT COMMAND UNDERSTANDING USING GRAMMAR-CONSTRAINED LARGE LANGUAGE MODELS}

%%% Change these to your keywords.  Keywords are automatically printed at the end of the abstract.
%%% This command must come BEFORE the end of the abstract.
%%% If you don't want keywords, omit the \keyword{..} command.

%% Abstract should be no more than 250 words
\begin{abstract}
Human-robot collaboration in industrial settings requires precise and reliable communication to enhance operational efficiency. While Large Language Models (LLMs) understand general language, they often lack the domain-specific rigidity needed for safe and executable industrial commands. To address this gap, this paper introduces a novel grammar-constrained LLM that integrates a grammar-driven Natural Language Understanding (NLU) system with a fine-tuned LLM, which enables both conversational flexibility and the deterministic precision required in robotics. Our method employs a two-stage process. First, a fine-tuned LLM performs high-level contextual reasoning and parameter inference on natural language inputs. Second, a Structured Language Model (SLM) and a grammar-based canonicalizer constrain the LLM's output, forcing it into a standardized symbolic format composed of valid action frames and command elements. This process guarantees that generated commands are valid and structured in a robot-readable JSON format. A key feature of the proposed model is a validation and feedback loop. A grammar parser validates the output against a predefined list of executable robotic actions. If a command is invalid, the system automatically generates corrective prompts and re-engages the LLM. This iterative self-correction mechanism allows the model to recover from initial interpretation errors to improve system robustness. We evaluate our grammar-constrained hybrid model against two baselines: a fine-tuned API-based LLM and a standalone grammar-driven NLU model. Using the Human Robot Interaction Corpus (HuRIC) dataset, we demonstrate that the hybrid approach achieves superior command validity, which promotes safer and more effective industrial human-robot collaboration.
\end{abstract}

\date{Version \versionno, \today}%% You can modify this information as desired. 
							%% Putting \date{} will suppress any date.  
							%% If this command is omitted, date defaults to \today
							%% This command must come somewhere before \maketitle

\maketitle %% This command creates the author/title/abstract block. Essential!

%%%%%%%%%%%%%%%%%%%%%%%%%%%%%%%%%%%%%%%%%%%%%%%%%%%%%%%%%%%%%%%%%%%%%%%%%%%%%%%%%%%%%%%%%%%%%%%%%%%%%%%
%%%%%%%%%%%%%%%%%%%%%  End of fields to be completed. Now write! %%%%%%%%%%%%%%%%%%%%%%%%%%%%%%%%%%%%%%

\section{Introduction}

Recent advances in artificial intelligence (AI) have begun to reshape manufacturing engineering, particularly in how robotic systems adapt, plan, and interact. As an emerging technology, robots in manufacturing environments can operate in changing conditions and assist human operators with consistency and endurance. However, current robots still rely on rigid control structures and lack the flexibility to respond to dynamic human input \cite{1}. As a result, these robotics systems fall short of smart manufacturing demands, which emphasize human participation \cite{2}, adaptive decision-making and learning capabilities \cite{3}, and seamless integration between offline programming and online interaction \cite{4}.

The integration of AI enables dynamic human-robot collaboration (HRC) through natural language communication, semantic understanding, and adaptive task planning. Specifically, Large Language Models (LLMs) embedded in robotic systems provide capabilities such as natural language understanding and task generation which enhances the interaction experience between humans and robots to create a high level of human centricity, generalization, and seamlessness \cite{5}. Recent studies have explored the use of LLMs in social contexts, showing that LLM-driven robots can generate situated task plans and interpret complex instructions within dynamic social environment \cite{6, 7}. While LLMs can be extended to engineering field, current research lacks optimization of LLMs for manufacturing constraints. Different from social scenarios, industrial environments require intuitive human-robot collaboration along with safety assurance for high-precision equipment \cite{8}.  

Within the specialty of manufacturing workspaces, this study aims to analyze the integration of LLMs into industrial robotic systems by comparing multiple LLMs approaches and evaluating their respective strengths and weaknesses. The rest of this paper is organized as follows: Section 2 introduce and compare existing LLMs methodologies. Section 3 includes the methodologies to execute the experiment. Section 4 shows the experiment results and possible improvements for the experiment. Section 5 is the conclusion of the paper.

%%%%%%%%%%%%%%%%%%%%%%%%%%%%%%%%%%%%%%%%%%%%%%%%%%%%%%%%%%%%%%%%%%%%%%

\section{Related Work}

LLMs can generally be categorized into three main architectural types: encoder-only, encoder-decoder, and decoder-only models \cite{5}. Encoder-only architectures contain encoder components and are used for understanding and representation tasks rather than text generation \cite{5}. Encoder-decoder models can encode the input to feature information and pass these to a decoder which generates outputs based on the encoded sequence. This design enables the capability in tasks requiring tight coupling between input and output sequences, such as translation \cite{9}. Decoder-only models contain decoder components and generate predictions based on prior tokens and context \cite{10}. The decoder-only architecture is the current dominant architecture \cite{11} including GPT-4 systems \cite{5}.

\subsection{Model Adaptation Methods}

To adapt general-purpose LLM to domain-specific tasks, prompt engineering and fine-tuning methods are widely used. Prompt-based approaches modify input instructions to guide reasoning behavior without changing model parameters \cite{12}, while parameter-efficient fine-tuning techniques enable efficient adaptation using limited domain datasets \cite{13}. Low-Rank Adaptation (LoRA) introduces trainable low-rank matrices into transformer layers, allowing efficient domain transfer while preserving the base model’s linguistic capability \cite{14}. These adaptation strategies are particularly useful in industrial and robotic environments where training resources or task-specific datasets may be limited.

\subsection{Decoding Control Strategies}

Decoding strategies determine how LLM outputs are generated during inference. Token-based autoregressive decoding predicts the next token sequentially, enabling flexible language generation across diverse domains \cite{15}. While prompt-guided generation can improve reasoning performance \cite{12}, token-level sampling relies on soft constraints and cannot guarantee structural validity of generated outputs \cite{16}.

Grammar-constrained decoding methods address this limitation by enforcing structural constraints during generation. These approaches integrate finite-state automata, probabilistic parsing, or external constraint modules to mask invalid tokens and ensure syntactic correctness \cite{17, 18}. Guided or soft-constrained sampling methods relax strict grammar enforcement while maintaining output coherence \cite{19, 20}. Recent constrained decoding frameworks dynamically construct token-validity masks using context-free grammar parsing states, enabling reliable structured generation for tasks such as code synthesis and robotic command interpretation \cite{16}. Systems such as XGrammar further optimize grammar execution through persistent parsing stacks and token pre-validation to reduce runtime overhead while preserving constraint guarantees \cite{21}. GRAMMAR-LLM integrates formal grammar classes directly into the decoding process to enforce syntactic correctness in linear time \cite{22}.

\subsection{LLM Deployment}

LLMs can be deployed either locally or through API-based service interfaces. API-based deployment provides scalable access to powerful foundation models such as GPT-4 and Llama through standardized interfaces \cite{15}. However, these services often limit low-level decoding control and direct parameter modification, requiring prompt engineering or preprocessing techniques to guide model behavior \cite{13}. In contrast, locally deployed open-source LLMs allow fine-tuning, latency control, and deeper integration with robotic execution systems, which are critical for safety-sensitive industrial applications.

%%%%%%%%%%%%%%%%%%%%%%%%%%%%%%%%%%%%%%%%%%%%%%%%%%%%%%%%%%%%%%%%%%%%%%%%%%%%%%%

\section{Methodology}

This section presents the three approaches evaluated in the study: An API-based LLM approach with or without Low-Rank Adaptation (LoRA) Fine Tuning; a grammar-based Natural Language Understanding (NLU) system that generate command through parsing rules; and the proposed hybrid grammar-based LLM, integrates the advantages of the API-based LLM and the Grammar-based NLU, and bridge the gap between open-domain conversational capabilities and the structured command understanding required for robotic execution.

\subsection{API-Based LLM}

\begin{figure}
\centering\includegraphics[width=0.7\linewidth]{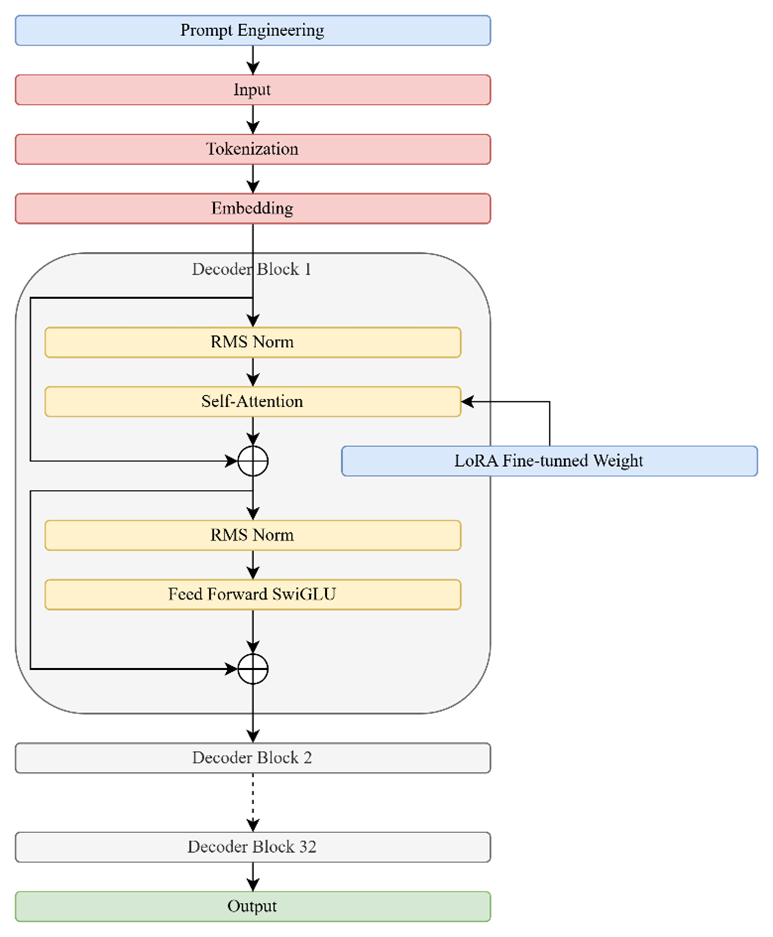}
\caption{META-LLAMA-3 Structure\label{fig:1}}
\end{figure}

The Meta-Llama-3-8B-Instruct \cite{23} model was selected due to its favorable balance between instruction-following capability and computational efficiency. The Llama-3 architecture employs Root Mean Square Layer Normalization (RMS Norm), which rescales hidden representations based on their root mean square value. Compared to traditional layer normalization, RMS Norm eliminates the mean-centering step and provides a simpler and computationally efficient alternative. This design enables the model to achieve competitive performance with fewer parameters, reduces memory overhead, and enables stable inference and LoRA fine-tuning on limited hardware. The 8-billion-parameter configuration can be efficiently deployed and fine-tuned on a single Graphics Processing Unit (GPU), while larger models would require significantly higher memory and distributed training resources. The method in this study integrates the model’s open-source API and local inference through a Python pipeline. To adapt the LLM to industrial manufacturing contexts, two strategies are applied: (1) prompt engineering which structures task related instructions and contextual cues to guild the LLM model; and (2) LoRA) \cite{14} fine-tuning which updates a small set of low-rank trainable parameters to integrate domain knowledge for improving semantic precision. These train parameters change how Self-Attention is computed. It changes the weight about how queries (Q), keys (K) and values (V) are projected, where Q search for relevant tokens, K represent token features, and V carry the aggregated information, while keeping the original pretrained weights. Data feeds to Feed Forward Swish-Gated Linear Unit (SwiGLU) to enhance the model’s representational capacity. Meta-Llama3-8b-Instruct model with prompt engineering and LoRA fine-tuning is shown in Figure~\ref{fig:1}.

\subsubsection{Prompt Engineering}

\begin{figure}
\centering\includegraphics[width=0.7\linewidth]{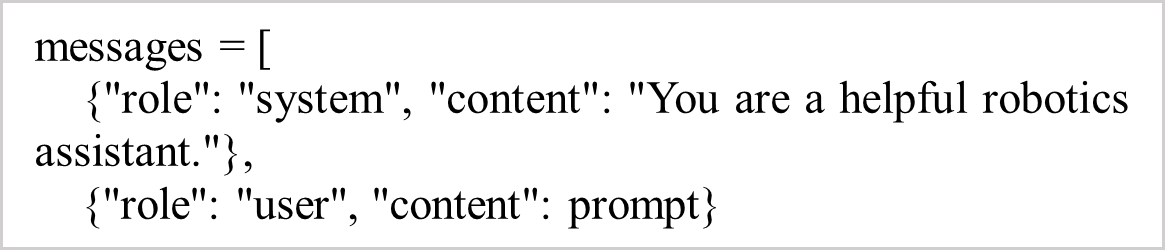}
\caption{Template of Two-role Chat and Prompt\label{fig:2}}
\end{figure}

Prompt engineering forms the external control layer for the LLMs. In this study, the Python pipeline implements a two-role chat template that establishes a compact prompt hierarchy: (1) System role declaration aims to define what the model is and how it should behave through interaction. For example, the prompt “You are a helpful robotics assistant.”, defines the model’s identity and behavior baseline. This ensures the models focus on instruction-following, safety, and task-oriented reasoning to avoid casual conversation or irrelevant details. This declaration steers its response towards instruction-oriented responses rather than conversational elaboration. (2) User role declaration provides the natural language input that the model needs to interpret. The role forces the model to extract semantic meaning from the input sentence and transfer it into structured frames. Sample input should be a command like “take the laptop and the book on the table on the couch.” Figure~\ref{fig:2} shows the prompt template with role definition. The role definitions let the LLM distinguish user input and the prompt given, and this simple prompt define the LLM characteristic to solve user provided input.

A tokenizer method serializes both roles into a single input sequence and ensures each query begins with the same semantic anchor. This consistency reduces response drift in different inputs.

Four model hyperparameters were tuned to balance syntactic precision and runtime stability. First, max\_new\_tokens: sets the maximum token count generated per inference, which prevent overflow and maintain fast response. Second, temperature: controls randomness in sampling. Lower values yield more determinism and tuning it is helpful to balance stability with natural variation. Third, top\_p: controls the percentage of tokens within probability mass to form a controlled-stochastic decoding regime. Last, repetition\_penalty: penalizes tokens reuse to avoid repetitive frames or loops for preserving coherence.

\begin{figure}
\centering\includegraphics[width=0.7\linewidth]{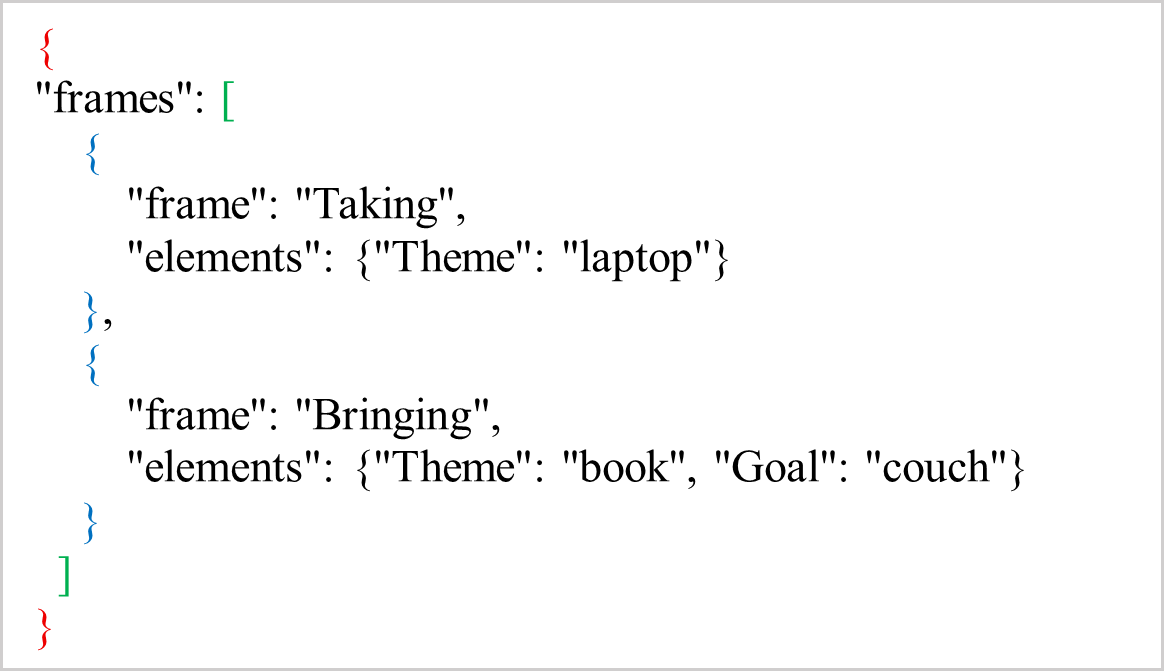}
\caption{Sample Output Structure\label{fig:3}}
\end{figure}

These hyperparameters produce a precise and parable structured output shown as follows, “frames” represents the sequence of actions extracted from the natural language input; “frame” denotes a standard action frame which correspond to executable robot actions; “elements” contain the semantic roles required by each action frame by their predefined structure. As an output example shown in Figure~\ref{fig:3}, the “frames” includes two actions in sequence, “frame”: “Taking” and “frame”: “Bringing”. Inside “frame”: “Taking”, “elements” indicate which item should be grasp. This “frame” can be understanded as “grasp the laptop.” The second action frame contains similar structure but with more semantic roles, indicate the action is bringing the book and delivering it to the couch.

\subsubsection{Fine-tuning Based on LoRA}

Fine-tuning component applies Low-Rank Adaptation (LoRA) \cite{14} to the Llama-3. This method provides efficient domain adaptation without full parameter retraining. LoRA decomposes the weight-update matrix $\Delta$W into two smaller matrices A and B of rank r. The approximation symbol is used because LoRA does not reconstruct the full update matrix exactly. It approximates $\Delta$W by low-rank product AB\^T that captures the dominant directions of change. The approximation function as shown below:

\begin{equation}
\label{equation 1}
    \Delta W \approx AB^T, A \in \mathbb{R}^{d \times r}, B \in \mathbb{R}^{r \times k} 
\end{equation}

Here, d represents the input dimension that is the feature size of the input vector. k represents the output dimension which is the number of projection outputs from the layer. r denotes the rank of the decomposition which controls how many parameters are trained in LoRA. During inference, the adapted weight becomes:

\begin{equation}
\label{equation 2}
    W' = W + \alpha \frac{AB^T}{r} 
\end{equation}

Where $\alpha$ is a scaling factor that controls the degree of influence from the LoRA adaptation on the base model. This formula allows the model to integrate new knowledge while preserving linguistic capability.

\begin{figure}
\centering\includegraphics[width=0.7\linewidth]{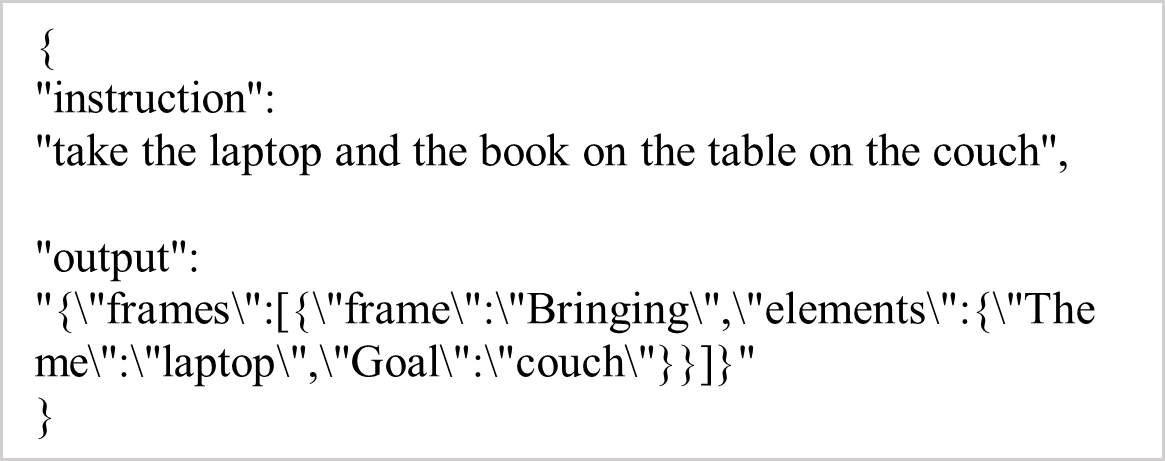}
\caption{Example of HURIC Dataset\label{fig:4}}
\end{figure}

The fine-tuning process utilizes the Human Robot Interaction Corpus (HuRIC) dataset \cite{24}. This dataset is an open-source dataset containing natural language robot commands annotated with corresponding robot action frames. Each data point is converted into JSONL format with fields of instructions and standard outputs. As shown in Figure~\ref{fig:4}, each data point after conversion includes the natural language instruction and a standard output. The dataset was divided into 90\% for training and 10\% for testing.

Training was executed through Parameter-Efficient Fine-Tuning (PEFT) \cite{25}, the general framework let LoRA be injected into the Llama-3. Accelerate library that manages hardware distribution and memory optimization \cite{26}. In addition to the rank r and scaling factor $\alpha$, several hyperparameters are tuned to balance accuracy and computational efficiency:

Dropout p: percentage of randomly disabled neurons, which sets low value to prevent over-regularization cost by HuRIC’s small data size. Sequence-length: maximum number of tokens per input with a safety margin. Batch-size: set based on GPU memory size. Learning-rate: control step size for updates to balance adaptation speed and stability.

\subsection{Grammar-Based NLU}

The Grammar-Based NLU relies on predefined syntactic rules that constrain allowable linguistic forms. Each rule maps a natural language input to a canonical frame representation composed of action frame types and semantic elements. This ensures each output conforms to the pre-defined action ontology. The Grammar-Based NLU consists of grammar parser, structured output, and validation.

\subsubsection{Grammar Parsing Architecture}

The parser includes three main components, including grammar definition, syntactic and semantic parsing, and schema construction.

The core grammar is implemented using Lark \cite{27}, a parsing library in Python. It can process the grammar written in Extended Backus-Naur form (EBNF) style syntax \cite{28}, a standard notation for writing grammar. Lark automatically builds a parser based on the written grammar for further process. This grammar defines the allowable command structure or robot interaction. It contains description of how verb phrases, noun phrases and prepositional phrases combine to form the robot command.

Syntactic and semantic parsing is performed with Lark’s Earley parser to build an Abstract Syntax Tree (AST) from the input natural language. During parsing process, each node corresponds to a command component (i.e. frame, theme, goal). The parsing workflow involves tokenization, tree construction, and semantic mapping. Once the AST is generated, a converter module extracts the AST frame’s action name and slot-value pairs and then reconstructs the content to a similar structure shown Figure~\ref{fig:3}.

\subsubsection{Structured Output Validation}

To ensure the output comply with the defined schema, a validation loop is applied after the schema generation. The module verifies the generated structure with the grammar’s valid frame list and performs auto-correction if necessary. If an invalid or undefined frame is detected, the system reconstructs the command via fallback parsing. In additional, for detected synonyms that are defined in the grammar, the system subtitles them to the same frame name for future processing. This process guarantees the syntactic integrity of the robot control commands.

\subsection{Hybrid Grammar-Based LLM}

The proposed hybrid grammar-based LLM integrates the advantages of the API-based LLM and the Grammar-based NLU into a unified interpretation pipeline. Unlike prior grammar-constrained decoding approaches that apply structural rules directly during token generation, the proposed method introduces a staged semantic-structural processing architecture. In this design, the LLM first performs unconstrained semantic interpretation to preserve contextual flexibility using the API-based LLM with LoRA fine-tuning and prompt engineering, working as a semantic front-end for the flexible and context-aware reasoning. The grammar-based NLU working as a grammar-constrained canonicalization and validation layer, enforces deterministic command structure. This separation enables semantic reasoning and structural determinism to be optimized independently, improving command reliability while maintaining linguistic adaptability. This canonicalization and validation layer provides possible feedback when validation failed. A new feedback prompt with structural guidance and correction will implement for increasing the result accuracy.

\subsubsection{Architecture and Workflow}

\begin{figure}
\centering\includegraphics[width=0.7\linewidth]{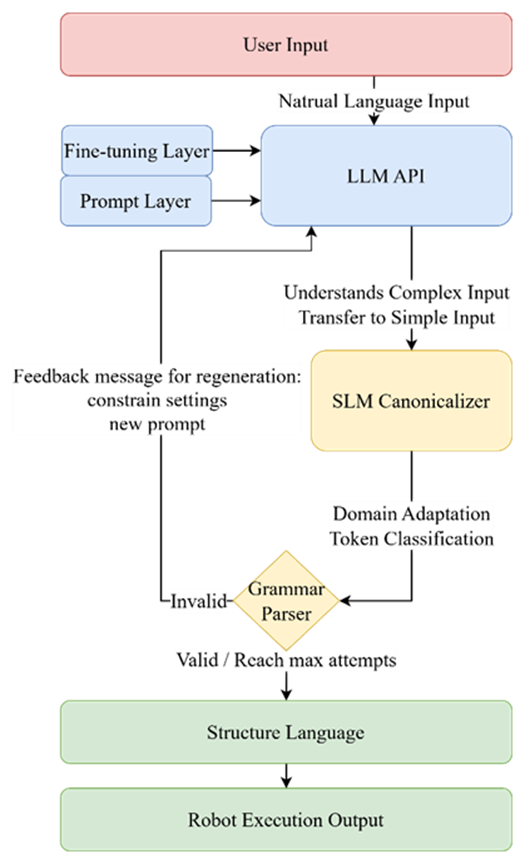}
\caption{Hybrid Grammar-based LLM Flow Chart\label{fig:5}}
\end{figure}

As shown in Figure~\ref{fig:5}, the Hybrid Grammar-based LLM integrates three main layers: semantic understanding, structural parsing and grammar validation, a feedback correction within a continuous loop.

Semantic understanding layer corresponds to the LLM API component and follows the same configuration as described in Section 3.1, a Meta-Llama-3-8B-Instruct model. This layer performs the semantic disambiguation, context inference, and command normalization, producing a raw structured output from the natural-language instruction without applying any structural constraints. 
Structural parsing and validation layer take the normalized output from the semantic understanding layer and pass it into the canonicalizer and grammar parser, which construct the AST and transforms it into a canonical JSON schema. The grammar output is further validated using a rule set defined in canonicalizer. 

An important component of this stage is the ELEMENT\_RULES dictionary, which specifies the allowable semantic elements for each action frame. For example:

"Bringing": ["Theme", "Beneficiary", "Goal", "Agent", "Source", "Area", "Manner"]

For the action frame “Bringing” defines allowable elements corresponding to commands in which the robot transfers an object from one point to another. Each word inside the ELEMENT\_RULES is the allowable element for each action frame and determine its canonical element names. Meanwhile, an auxiliary mapping ELEMENT\_KEY\_REMAP remaps defined synonyms and surface variations to their canonical element names (e.g. remap “Containing\_portal” to “Containing\_object”). A separate optional structural filter can be modified by user to discard any unexpected or unsupported elements.

\begin{figure}
\centering\includegraphics[width=0.7\linewidth]{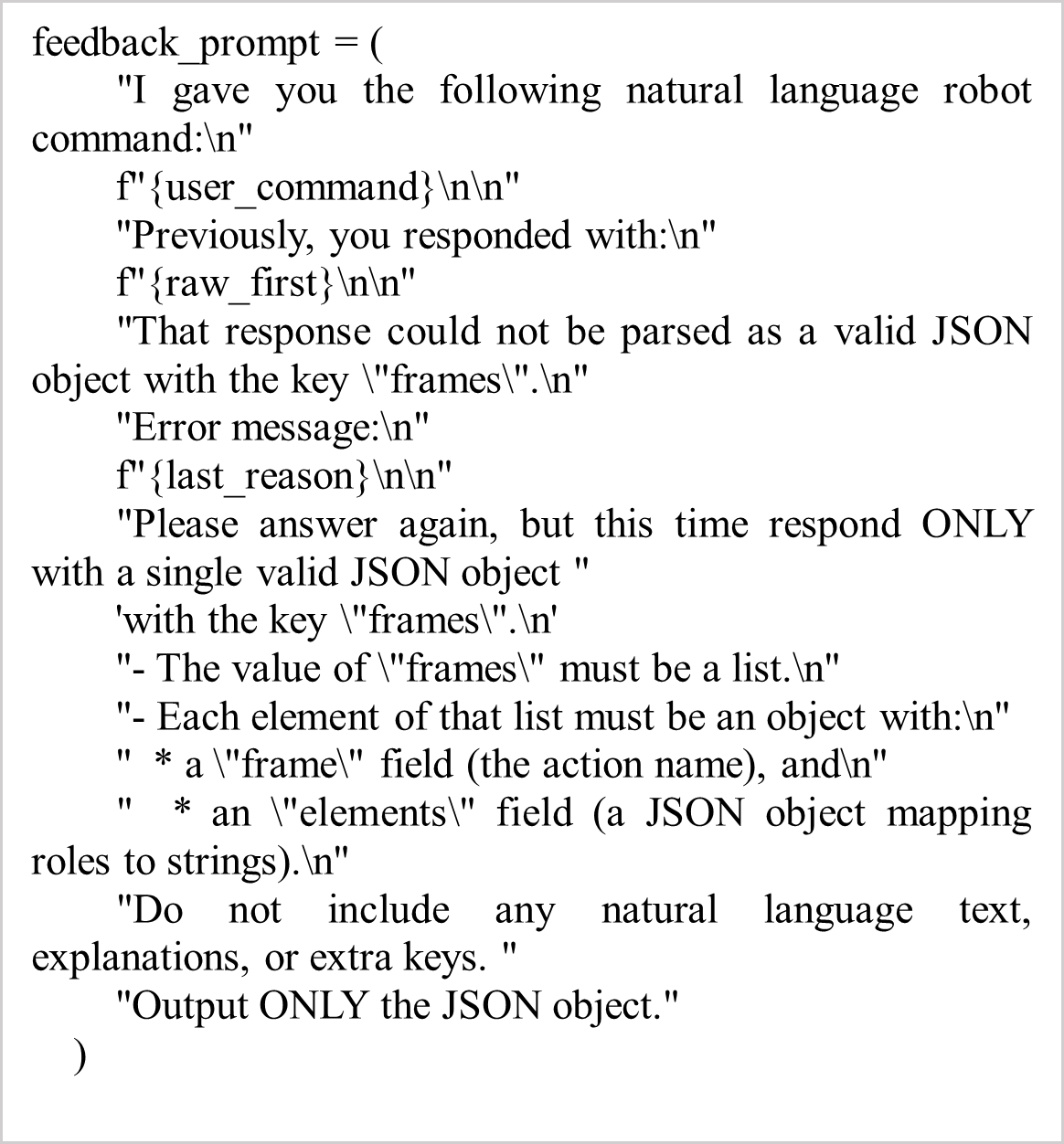}
\caption{Template of Feedback Prompt\label{fig:6}}
\end{figure}

A key novelty of the proposed architecture is the feedback-driven structural correction loop. Instead of rejecting invalid outputs, the system generates correction prompts base on the grammar parser’s result. The new generated prompt guide the LLM to optimize the output with the structurally inconsistent elements while preserving extracted valid semantic components. The correction loop begins with a light-constraint stage, where the system attempts to retain valid action frames and discard unexpected elements. If a fully valid structure cannot be recovered, a strict-constraint validation stage is applied to enforce compliance with predefined grammar rules and action lists. When invalid or missing information persists, the system automatically generates an error report and re-prompts the LLM with corrective guidance indicating the unsupported elements or actions. To prevent infinite iterations, the number of correction attempts is limited. Once the final structure passes validation, the JSON command is forwarded to the execution layer.  The feedback prompt template is shown in Figure~\ref{fig:6}, where user input, previous generated response, error message from the grammar parser, and defined action frame will be given to LLM for regeneration. This iterative correction mechanism improves structural reliability while maintaining semantic flexibility, enabling stable convergence toward executable command representations.

%%%%%%%%%%%%%%%%%%%%%%%%%%%%%%%%%%%%%%%%%%%%%%%%%%%%%%%%%%%%%%%%%%%%%%%%%%%%%%%

\section{Results and Discussion}

\subsection{System Configuration}

Experiments in this study are conducted on a local workstation running Ubuntu 24.04.3 LTS under the Windows 11 Pro WSL 2 environment. The system is equipped with an AMD Ryzen 7 5800X CPU, 32 GB DDR4 RAM, and an NVIDIA RTX 3080 GPU featuring 10 GB of VRAM. Each proposed model is executed in a Python 3.11 virtual environment using PyTorch 2.3 and the Hugging Face Transformers library, with CUDA acceleration enabled within WSL.

\subsection{Results}

Experiments are evaluated using the HuRIC dataset \cite{22}, which contains natural-language instructions and corresponding structured command annotations for robotic tasks. The dataset containing total of 656 data covering 16 distinct action frames, and each frame is associated with a predefined set of element types. The dataset includes several sentence structures: imperative command, interrogatives, and polite requests. The dataset is randomly divided into 90\% for training and 10\% for testing. All quantitative results reported below are obtained from the testing set, which contains a total of 66 test samples. 

Each predictive output is compared against the HuRIC-provided reference commands to measure the JSON Exact Match (EM), a metric that measures whether the predicted action frame and its semantic elements exactly match the reference annotation.

To capture partial correctness, JSON Similarity is also calculated. This metric quantifies the structural and semantic overlap between the predicted and reference JSON objects by comparing their frames, keys, and values. JSON Similarity is computed using a Jaccard-index-based formula:

\begin{equation}
\label{equation 3}
    JSON Similarity = \frac{|A \cap B|}{|A \cup B|}
\end{equation}

Where A and B are sets of key–value pairs (e.g., “frame=Bringing”, “Goal=kitchen”) from the prediction and reference, respectively.

\subsubsection{API-based LLM w/o Fine-tuning Results}

Within the all the testing sets, this model achieved an Exact Match (EM) rate of 0\%, as there is no exact match with the reference annotation from the dataset. The model has a 3.03\% success output rate; the successful output results have 0\% JSON Similarity, as the output elements were non important words captured from the input natural language with no intersection with the reference annotation. Shown this approach lack of ability to process natural language to robot command.

\subsubsection{API-based LLM w/ Fine-tuning Results}

\begin{figure}
\centering\includegraphics[width=0.7\linewidth]{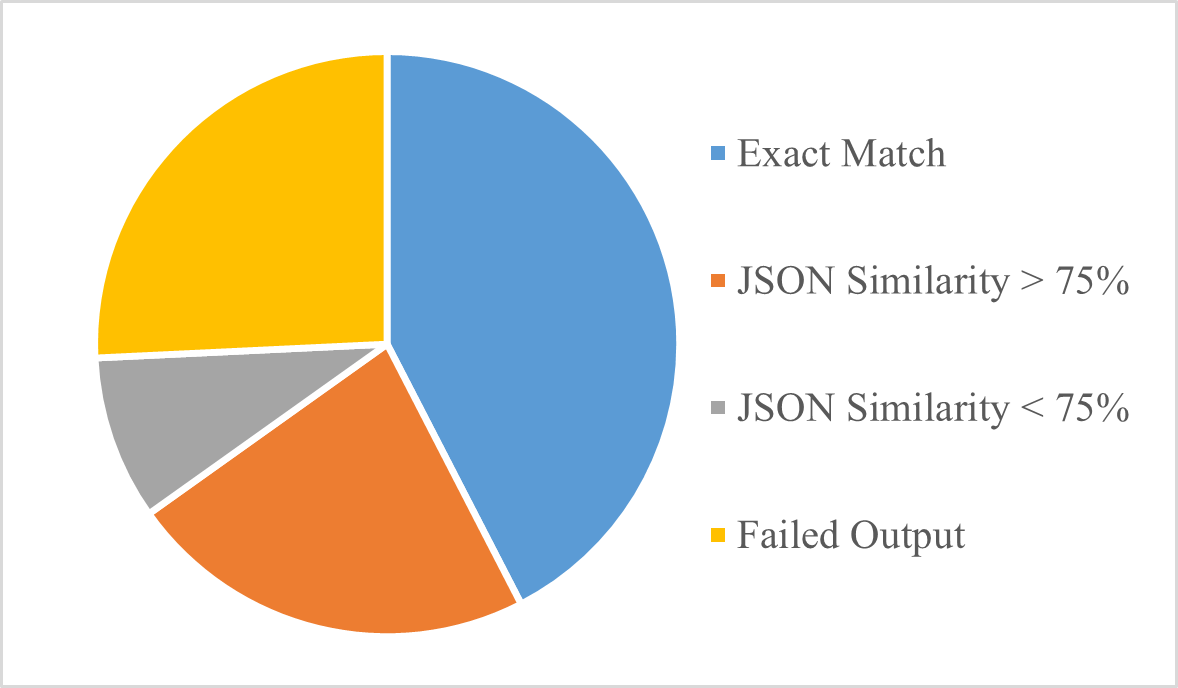}
\caption{Percentage of API-based LLM w/ Fine-tuning Result\label{fig:7}}
\end{figure}

As shown in Figure~\ref{fig:7}, this model achieved an Exact Match (EM) rate of 36.4\% and an average JSON Similarity of 72.3\%. Among outputs that did not achieve an exact match, 71.3\% reached at least 75\% JSON Similarity, which indicate the LLM successfully extract most of needed token from the natural language input. These results demonstrate the model’s moderate structural accuracy and reasonable semantic understanding.

The model also has 25.76\% of responses failed to generate a valid JSON structure, because the API-based LLM tends to reply in conversation model instead of structured output.

\subsubsection{Grammar-based NLU Results}

\begin{figure}
\centering\includegraphics[width=0.7\linewidth]{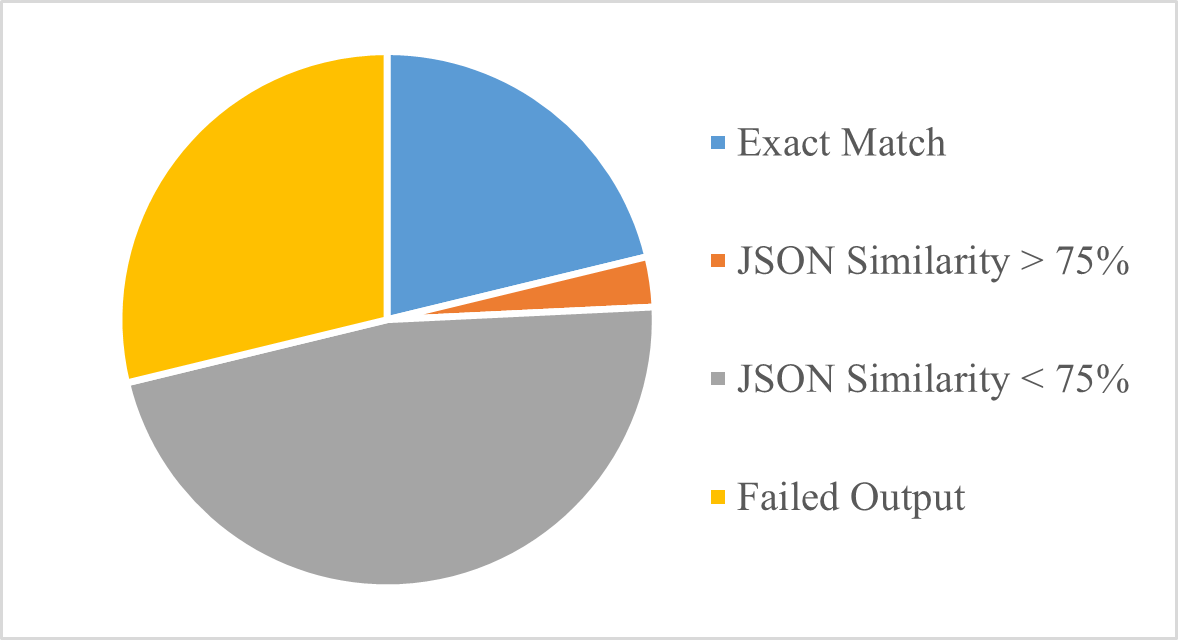}
\caption{Percentage of Grammar-based NLU Result\label{fig:8}}
\end{figure}

As shown in Figure~\ref{fig:8}, this model achieved an Exact Match (EM) rate of 21.21\% and an average JSON Similarity of 42.25\%. For success outputs but not EM results, only 6.06\% of them reach 75\% JSON Similarity, indicating though the NLU system extract the action frame but lack of ability to exact the elements from the natural language input.

The Grammar-based NLU model also has 28.79\% of responses failed to generate a valid JSON structure since unexpected token appears in the input.

\subsubsection{Hybrid Grammar-based LLM Results}

\begin{figure}
\centering\includegraphics[width=0.7\linewidth]{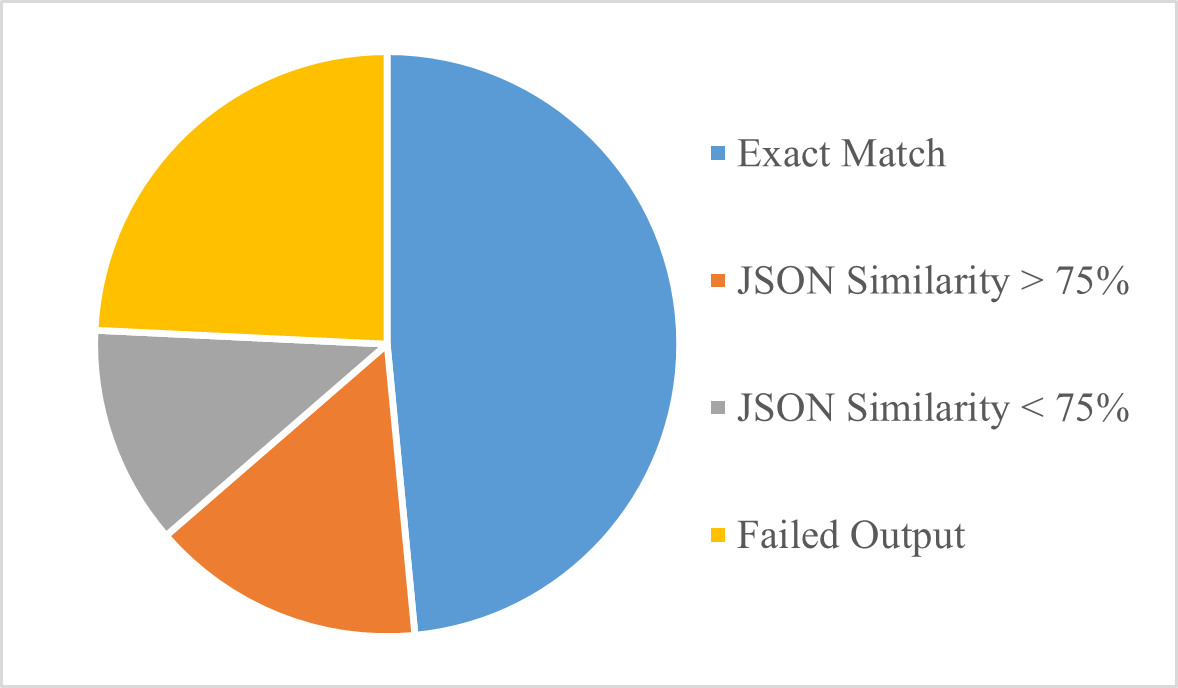}
\caption{Percentage of Hybrid Grammar-based LLM Result\label{fig:9}}
\end{figure}

As shown in Figure~\ref{fig:9}, this model achieved an Exact Match (EM) rate of 48.48\% and an average JSON Similarity of 88.07\%. For success outputs but not EM results, 61.11\% of them reach 75\% JSON Similarity, indicating the hybrid method’s improved performance.

The model also has 24.24\% of failed output. In this method, all failed outputs are being constrained to empty output “{"frames": []}”.

\subsection{Discussion}

TABLE 1 shows the comparison between the API-based LLM w/o Fine-tuning, API-based LLM w/ Fine-tuning, Grammar-based NLU, and the Hybrid Grammar-based LLM results in Exact Match and JSON Similarity.

In terms of overall successful output rate, the API-based LLM w/o Fine-tuning only have 3.03\% success output, but JSON Similarity of the output remain 0, shows its lack of ability to transfer natural language to robot command, other three methods show relatively close performance: 74.24\% for the API-based LLM, 71.21\% for the Grammar-based NLU, and 75.76\% for the Hybrid Grammar-based LLM. This similarity suggests that all systems can produce structurally valid outputs in most cases. Therefore, the performance gap observed across models is not primarily due to generation success, but rather in the quality and precision of the structured output, as reflected in the JSON Exact Match (EM) and JSON Similarity metrics.

Comparing with its successful output rate, API-based LLM achieved moderate precision of exact match. This indicates that the model can understand contextual meaning and generate appropriate action frames. However, it also causes a portion of the responses to appear as free-form texts rather than strict JSON element names. This issue comes from the model’s probabilistic decoding process, where the lack of structural constraints allows flexible token generation but increases formatting inconsistency. The correlation between moderate EM and high JSON Similarity reflects that the model captures semantic intent but lacks deterministic formatting control.

By contrast, the Grammar-based NLU system presents strong syntactic control but weaker semantic adaptability. Its high successful output rate indicated it is capable of extract the action frame from most of the input. The lowest number of JSON Similarity is due to the strict token expectation as the input may not show exact grammar as the defined grammar. Though the action may successfully exact, incomplete semantic details may capture by following the defined grammar rule.

\begin{figure}
\centering\includegraphics[width=0.7\linewidth]{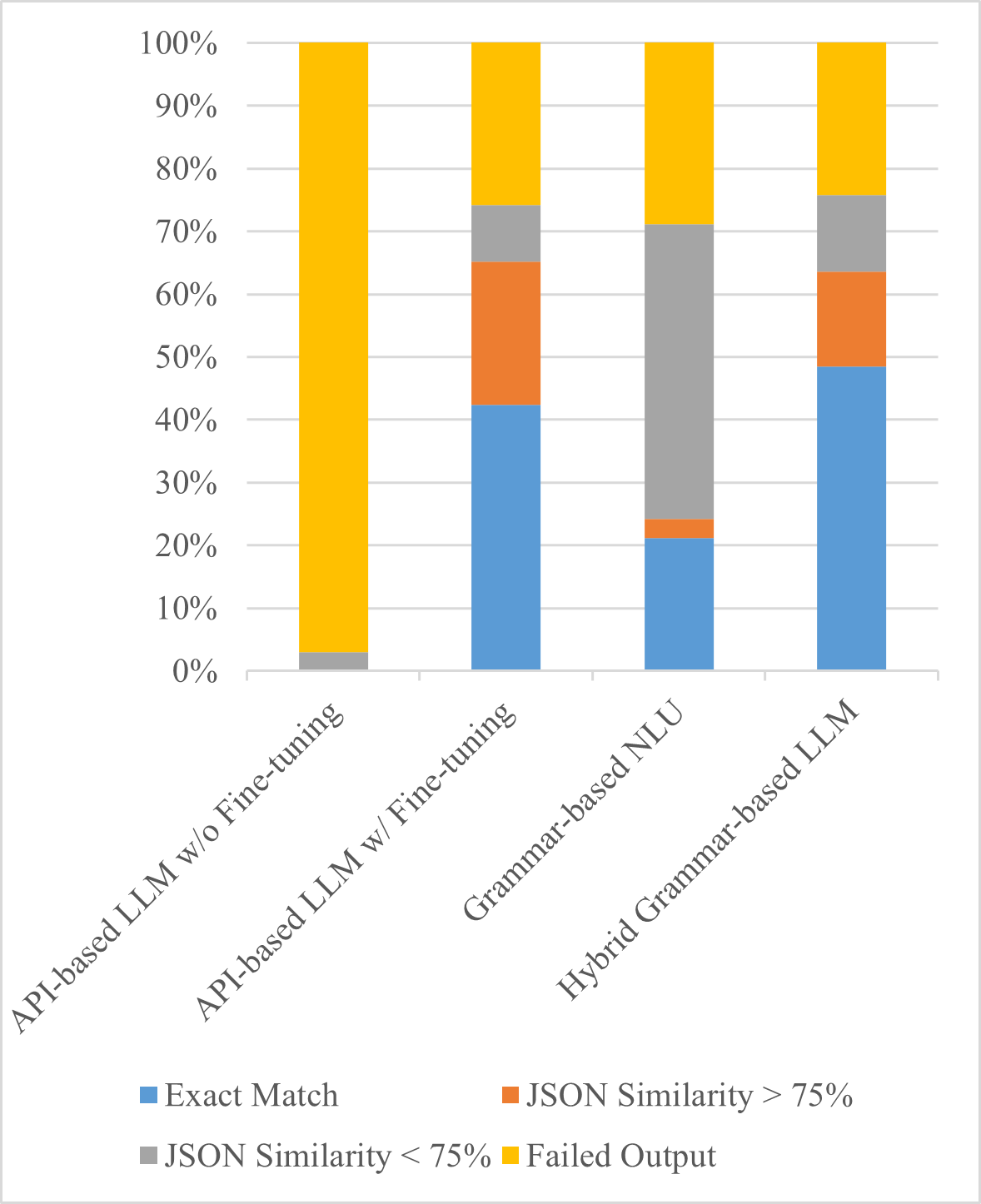}
\caption{Output Distribution of Four Methods\label{fig:10}}
\end{figure}

The Hybrid Grammar-based LLM combines the flexibility of the API-based model and the structure enforcement of the Grammar-based NLU system. As shown in Figure~\ref{fig:10}, it reaches both the highest Exact Match rate and the lowest failed output within these three methods.  Although the number of JSON Similarity < 75\% increases, it still has the highest overall JSON Similarity. This shows its outputs’ semantically complete and structurally consistent, balances API-based LLM prioritizes semantic fluency and the Grammar-based NLU prioritizes rule integrity.

\begin{figure}
\centering\includegraphics[width=0.7\linewidth]{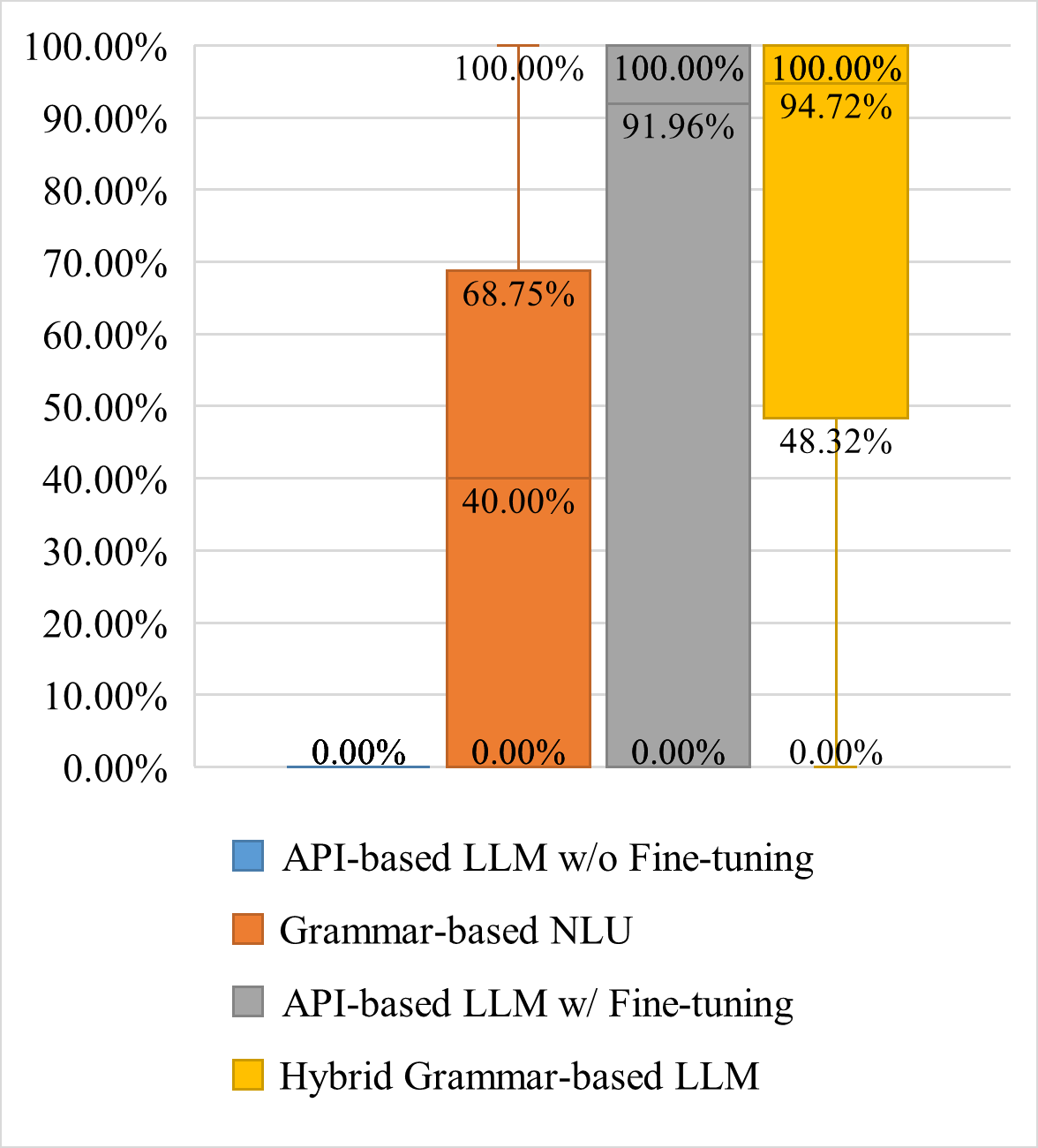}
\caption{JSON Similarity Box Plot of Four Methods\label{fig:11}}
\end{figure}

As shown in Figure~\ref{fig:11}, the lower 25\% JSON Similarity is 48\% for the Hybrid Grammar-based LLM, representing a better successful output rate than both Grammar-based NLU and API-based LLM’s lower 25\% JSON Similarity of 0\%. The median of Hybrid Grammar-based LLM reach 95\% JSON Similarity shows improving in accuracy than API-based LLM’s 92\% and Grammar-based NLU’s 40\%.

\begin{table}[t]
\caption{Output Comparison of Four Methods}
\label{tab:1}
\centering
\resizebox{\columnwidth}{!}{
\begin{tabular}{lcccc}
\toprule
Experiment & Success Output & Exact Match & JSON Similarity & Failed Output \\
\midrule
API-based LLM w/o Fine-tuning & 3.03 & 0.00 & 0.00 & 96.67 \\
API-based LLM w/ Fine-tuning & 74.24 & 36.3 & 72.3 & 25.76 \\
Grammar-based NLU & 71.21 & 21.21 & 42.55 & 28.79 \\
Hybrid Grammar-based LLM & 75.76 & 48.48 & 88.07 & 24.24 \\
\bottomrule
\end{tabular}
}
\end{table}

%%%%%%%%%%%%%%%%%%%%%%%%%%%%%%%%%%%%%%%%%%%%%%%%%%%%%%%%%%%%%%%%%%%%%%%%%%%%%%%

\section{Conclusion}

This paper presented a Hybrid Grammar-based LLM framework that is integral with API-based LLM with grammar-based constraint. The results demonstrate the developed method improves in its structural consistency and semantic accuracy compared with individual API-based LLM and grammar-based NLU approaches.

Several limitations are concluded here to guide future studies. The first limitation concerns the dataset scale and linguistic composition. The HuRIC dataset used in this study contains only a limited number of samples (66 for testing, 590 for training). These also include interrogative and polite request forms that can lead the LLM to interpret them as conversation rather than instruction. Potential solutions include expanding the dataset to imperative and structured instructions. An additional layer to pre-process the input natural language can also substile those interrogative and request forms to a new input for further process.

The second limitation is in grammar constraints in the Hybrid Grammar-based approach. Grammar rules built for each sentence yield higher accuracy (71.21\%) because they are optimized for that structure. When all rules are merged into unified grammar, the number of possible elements per action frame increases. This introduces structural ambiguity and reduces exact-match performance. One improvement is hierarchical grammar design, where general frame templates are defined at the top level and element rules are modularized by action type. Another option is probabilistic grammar weighting or adaptive parsing, which would allow the system to prioritize the most likely frame and element combinations.

A further limitation is the reliance on a single base API model (Meta-Llama-3-8B-Instruct). Its stylistic biases and decoding preferences affect both output format and structural fidelity. Future work should examine alternative foundation models (e.g., GPT-4) and compare them under identical grammar-constrained conditions.

For near future, creating an industrial specific robotic dataset will benefit in this research. As contain frequent used natural language command in industrial environment, the dataset will be easy to develop related robot action list for better grammar constraints, and avoid interrogative and polite request forms natural languages. The dataset could also provide multiple actions included commands to training possible hierarchical grammar design, probabilistic grammar weighting, or adaptive parsing approach could benefits in complex natural language understanding. Change the base API model will also be conducted to further validate on the Hybrid Grammar-based LLM approach. Grammar-parser will have same constraints to check the improve of JSON Similarity could reproduced in other base LLM model.

%%%%%%%%%%%%%%%%%%%%%%%%%%%%%%%%%%%%%%%%%%%%%%%%%%%%%%%%%%%%%%%%%%%%%%

%%%%%%%%%%%%%%%%%%%%%%%%%%%%%%%%%%%%%%%%%%%%%%%%%%%%%%%%%%%%%%%%%%%%%%
\section*{Acknowledgment} %% ASME requests this exact spelling, singular.

Acknowledge individuals, institutions, or companies that supported the authors in preparing the work. Those mentioned might have provided technical support, insightful comments or conversations, materials used in the work, or access to facilities.

%%%%%%%%%%%%%%%%%%%%%%%%%%%%%%%%%%%%%%%%%%%%%%%%%%%%%%%%%%%%%%%%%%%%%%

%%%%%%%%%%%%%  BIBLIOGRAPHY  %%%%%%%%%%%%%%%%%%%%%%%%%%%%%%%%%%%%%%%%%

\nocite{*} %% <=== delete this line - unless you wish to typeset the entire contents of your .bib file.

\bibliographystyle{asmejour}   %% .bst file that follows ASME journal format. Do not change.

\bibliography{MSEC_Conference_Paper_Reference} %% <=== change this to name of your bib file

\begin{thebibliography}{10}
\newcommand{\enquote}[1]{``#1''}
\providecommand{\url}[1]{\texttt{#1}}
\providecommand{\urlprefix}{}
\expandafter\ifx\csname urlstyle\endcsname\relax
  \providecommand{\doi}[1]{doi:\discretionary{}{}{}#1}\else
  \providecommand{\doi}{doi:\discretionary{}{}{}\begingroup \urlstyle{rm}\Url}\fi
\providecommand{\eprint}[2][]{\urlprefix\url{#1#2}}
\providecommand{\hrefurl}[2][]{\href{#1}{#2}}

\bibitem{1}
Buerkle, A., Eaton, W., Al-Yacoub, A., Zimmer, M., Kinnell, P., Henshaw, M., Coombes, M., Chen, W.-H., and Lohse, N., 2023, \enquote{Towards Industrial Robots as a Service (IRaaS): Flexibility, Usability, Safety and Business Models,} \hrefurl{https://doi.org/10.1016/j.rcim.2022.102484}{Robotics and Computer-Integrated Manufacturing}, \textbf{81}, p. 102484.

\bibitem{2}
Lu, Y., Zheng, H., Chand, S., Xia, W., Liu, Z., Xu, X., Wang, L., Qin, Z., and Bao, J., 2022, \enquote{Outlook on Human-Centric Manufacturing Towards Industry 5.0,} Journal of Manufacturing Systems, \textbf{62}, pp. 612--627.

\bibitem{3}
Wang, L., 2019, \enquote{From Intelligence Science to Intelligent Manufacturing,} Engineering, \textbf{5}(4), pp. 615--618.

\bibitem{4}
Angleraud, A., Ekrekli, A., Samarawickrama, K., Sharma, G., and Pieters, R., 2024, \enquote{Sensor-Based Human--Robot Collaboration for Industrial Tasks,} Robotics and Computer-Integrated Manufacturing, \textbf{86}, p. 102663.

\bibitem{5}
Dong, W., Li, S., and Zheng, P., 2025, \enquote{Toward Embodied Intelligence-Enabled Human--Robot Symbiotic Manufacturing: A Large Language Model-Based Perspective,} \hrefurl{https://doi.org/10.1115/1.4068235}{Journal of Computing and Information Science in Engineering}, \textbf{25}(5), p. 050801.

\bibitem{6}
Yang, G., Huang, X., and Guo, Y., 2024, \enquote{Semantic Map Based Robot Navigation with Natural Language Input,} \textit{2024 33rd IEEE International Conference on Robot and Human Interactive Communication (ROMAN)}, Pasadena, CA, USA, pp. 1689--1696.

\bibitem{7}
Singh, I., Blukis, V., Mousavian, A., Goyal, A., Xu, D., Tremblay, J., Fox, D., Thomason, J., and Garg, A., 2023, \enquote{ProgPrompt: Generating Situated Robot Task Plans Using Large Language Models,} \textit{Proceedings of the 2023 IEEE International Conference on Robotics and Automation (ICRA)}, pp. 11523--11530.

\bibitem{8}
Rodr{\'i}guez-Guerra, D., Sorrosal, G., Cabanes, I., and Calleja, C., 2021, \enquote{Human-Robot Interaction Review: Challenges and Solutions for Modern Industrial Environments,} IEEE Access, \textbf{9}, pp. 108557--108578.

\bibitem{9}
Ott, M., Edunov, S., Baevski, A., Fan, A., Gross, S., Ng, N., Grangier, D., and Auli, M., 2019, \enquote{fairseq: A Fast, Extensible Toolkit for Sequence Modeling,} \textit{NAACL-HLT 2019}, Minneapolis, MN.

\bibitem{10}
Wang, T., Roberts, A., Hesslow, D., Le~Scao, T., Chung, H.~W., Beltagy, I., Launay, J., and Raffel, C., 2022, \enquote{What Language Model Architecture and Pretraining Objective Works Best for Zero-Shot Generalization?} \textit{International Conference on Machine Learning}, PMLR, Baltimore, MD, pp. 22964--22984.

\bibitem{11}
Qorib, M.~R., Moon, G., and Ng, H.~T., 2024, \enquote{Are Decoder-Only Language Models Better Than Encoder-Only Language Models in Understanding Word Meaning?} \textit{Findings of the Association for Computational Linguistics: ACL 2024}, Bangkok, Thailand, pp. 16339--16347.

\bibitem{12}
Gereti, M., Robinson, A., Williams, S., et~al., 2024, \enquote{Token-Based Prompt Manipulation for Automated Large Language Model Evaluation,} TechRxiv.

\bibitem{13}
Lee, D., Lee, J., and Shin, D., 2024, \enquote{GPT Prompt Engineering for a Large Language Model-Based Process Improvement Generation System,} Korean Journal of Chemical Engineering, \textbf{41}, pp. 3263--3286.

\bibitem{14}
Hu, E.~J., Shen, Y., Wallis, P., Allen-Zhu, Z., Li, Y., Wang, S., Wang, L., and Chen, W., 2022, \enquote{LoRA: Low-Rank Adaptation of Large Language Models,} \textit{International Conference on Learning Representations (ICLR)}.

\bibitem{15}
Vemprala, S., Bonatti, R., Bucker, A., and Kapoor, A., 2023, \enquote{ChatGPT for Robotics: Design Principles and Model Abilities,} arXiv preprint arXiv:2306.17582.

\bibitem{16}
Geng, S., Josifoski, M., Peyrard, M., and West, R., 2024, \enquote{Grammar-Constrained Decoding for Structured NLP Tasks without Finetuning,} arXiv preprint arXiv:2305.13971.

\bibitem{17}
Koo, T., Liu, F., and He, L., 2024, \enquote{Automata-Based Constraints for Language Model Decoding,} arXiv preprint arXiv:2407.08103.

\bibitem{18}
Park, K., Zhou, T., and D'Antoni, L., 2025, \enquote{Flexible and Efficient Grammar-Constrained Decoding,} arXiv preprint arXiv:2502.05111.

\bibitem{19}
Willard, B.~T. and Louf, R., 2023, \enquote{Efficient Guided Generation for Large Language Models,} arXiv preprint arXiv:2307.09702.

\bibitem{20}
Zhang, H., Dang, M., Peng, N., and Van~den Broeck, G., 2023, \enquote{Tractable Control for Autoregressive Language Generation,} \textit{Proceedings of the International Conference on Machine Learning (ICML)}, pp. 40932--40945.

\bibitem{21}
Dong, Y., Ruan, C.~F., Cai, Y., Lai, R., Xu, Z., Zhao, Y., and Chen, T., 2025, \enquote{XGrammar: Flexible and Efficient Structured Generation Engine for Large Language Models,} arXiv preprint arXiv:2411.15100.

\bibitem{22}
Tuccio, G., Bulla, L., Madonia, M., Gangemi, A., and Mongiovi, M., 2025, \enquote{GRAMMAR-LLM: Grammar-Constrained Natural Language Generation,} \textit{Findings of the Association for Computational Linguistics: ACL 2025}, pp. 3412--3422.

\bibitem{23}
Grattafiori, A., Dubey, A., Jauhri, A., Pandey, A., Kadian, A., Al-Dahle, A., Letman, A., Mathur, A., Schelten, A., Vaughan, A., et~al., 2024, \enquote{The Llama 3 Herd of Models,} arXiv preprint arXiv:2407.21783.

\bibitem{24}
Vanzo, A., Croce, D., Bastianelli, E., Basili, R., and Nardi, D., 2020, \enquote{Grounded Language Interpretation of Robotic Commands Through Structured Learning,} \hrefurl{https://doi.org/10.1016/j.artint.2019.103181}{Artificial Intelligence}, \textbf{278}, p. 103181.

\bibitem{25}
Mangrulkar, S., Gugger, S., Debut, L., Belkada, Y., Paul, S., and Bossan, B., 2022, \enquote{PEFT: State-of-the-Art Parameter-Efficient Fine-Tuning Methods,} GitHub Repository, \urlprefix\url{https://github.com/huggingface/peft}

\bibitem{26}
Gugger, S., Debut, L., Wolf, T., Schmid, P., Mueller, Z., Mangrulkar, S., Sun, M., and Bossan, B., 2022, \enquote{Accelerate: Training and Inference at Scale Made Simple, Efficient and Adaptable,} GitHub Repository, \urlprefix\url{https://github.com/huggingface/accelerate}

\bibitem{27}
Shinan, E., 2020, \enquote{Lark: A Modern Parsing Library for Python,} Lark Documentation, \urlprefix\url{https://lark-parser.readthedocs.io/en/stable/index.html}

\bibitem{28}
{International Organization for Standardization}, 1996, \enquote{{ISO/IEC 14977:1996 --- Information Technology --- Syntactic Metalanguage --- Extended BNF},} ISO, Geneva.

\end{thebibliography}

%%%%%%%%%%%%%%%%%%%%%%%%%%%%%%%%%%%%%%%%%%%%%%%%%%%%%%%%%%%%%%%%%%%%%%

%% To omit final list of figures and tables, use the class option [nolists]

\end{document}